%% file: SoftRobots.tex
\documentclass[letterpaper, 10 pt, conference]{ieeeconf}  

\IEEEoverridecommandlockouts                              
\usepackage{algorithm}
\usepackage{amsmath}
\usepackage{amssymb}
\usepackage{url}
\usepackage{hyperref}
\usepackage{bm}
\usepackage{booktabs}
\usepackage{calc}
\usepackage{caption}
\usepackage{subcaption}
\usepackage[table]{xcolor} 
\usepackage{enumerate}
\usepackage{graphicx}
\usepackage{multicol}
\usepackage{siunitx}
\usepackage{cite}
\input{defs}

\title{\LARGE \bf
Shape-Space Graphs: \\
Fast and Collision-Free Path Planning for Soft Robots
}

\author{Carina Veil$^{1}$, Moritz Flaschel$^{2}$, and Ellen Kuhl$^{1}$
	\thanks{This work was supported by the NSF CMMI Award 2318188 Mechanics of Bioinspired Soft Slender Actuators and the ERC Advanced Grant 101141626 DISCOVER. Corresponding author: Carina Veil.}
	\thanks{$^{1}$C. Veil and E. Kuhl are with the Department of Mechanical Engineering, Stanford University, Stanford, CA 94305, USA. 
    {\tt\small \{cveil,ekuhl\}@stanford.edu}}
   \thanks{$^{2}$ M. Flaschel is with the Institute of Applied Mechanics, Friedrich-Alexander-Universität Erlangen–Nürnberg, 91058 Erlangen, Germany.
		{\tt\small moritz.flaschel@fau.de}
}}

\begin{document}

\maketitle
\thispagestyle{empty}
\pagestyle{empty}

\begin{abstract}
Soft robots, inspired by elephant trunks or octopus arms, offer extraordinary flexibility to bend, twist, and elongate in ways that rigid robots cannot.
However, their motion planning remains a challenge, especially in cluttered environments with obstacles, due to their highly nonlinear and infinite-dimensional kinematics.
Here, we present a graph-based path planning tool for an elephant-trunk-inspired soft robot designed with three artificial muscle fibers that allow for continuous deformation through contraction.
Using a biomechanical model \rev{that integrates} morphoelastic and active filament theories, we precompute a shape library and construct a $k$-nearest neighbor graph in \emph{shape space}, ensuring that each node corresponds to a valid robot shape.
For the graph, we use signed distance functions to prune nodes and edges colliding with obstacles, and define multi-objective edge costs based on geometric distance and actuation effort, enabling energy-\rev{aware} planning with collision avoidance.
We demonstrate that our algorithm reliably avoids obstacles and generates feasible paths within milliseconds from precomputed graphs using Dijkstra's algorithm. 
We show that including energy costs can drastically reduce the actuation effort compared to geometry-only planning, at the expense of longer tip trajectories. 
Our results highlight the potential of shape-space graph search for fast and reliable path planning in the field of soft robotics, paving the way for real-time applications in surgical, industrial, and assistive settings.
\end{abstract}



\begin{figure*}
    \centering
    \includegraphics[width=\textwidth]{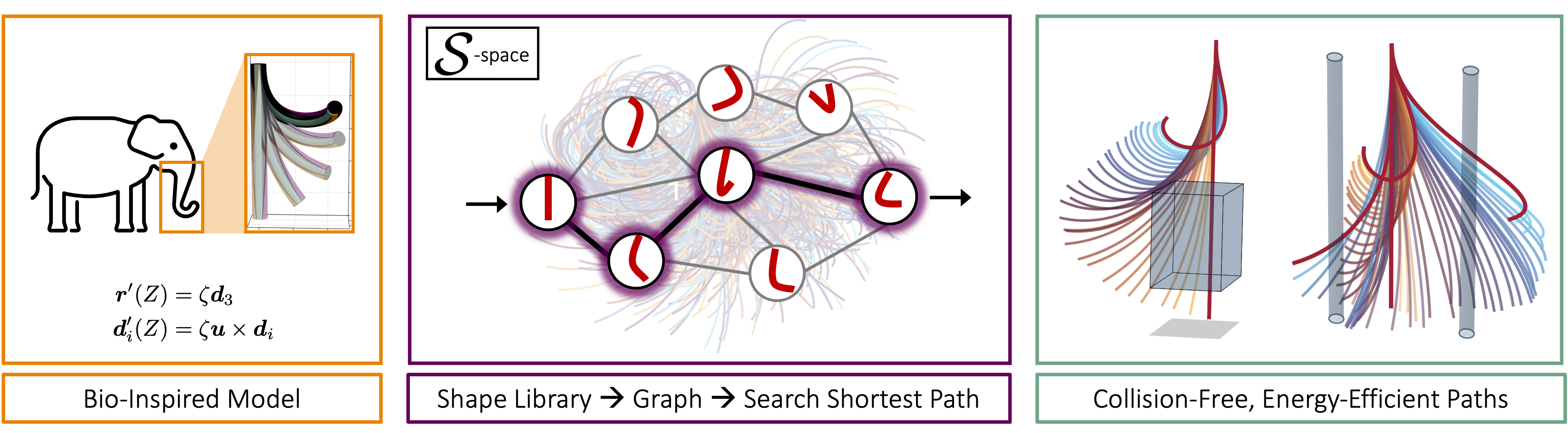}
    \caption{\textbf{Shape-space graphs}: We use a biomechanical model of our three-fiber soft robotic arm inspired by the elephant’s muscular structure to precompute a large shape library paired with corresponding fiber activation tuples. Constructing a $k$-nearest neighbor graph in the robot's \emph{shape space} (\shs{}) transforms the path planning problem into a graph search, enabling efficient multi-objective optimization that minimizes activation effort while avoiding obstacles.}
\end{figure*}

\section{Introduction}

Soft continuum robots, inspired by biological structures such as elephant trunks, octopus arms, or plant stems, can bend, twist, and elongate in ways that rigid robots cannot. This extreme flexibility enables them to navigate tight spaces, adapt to complex environments, and interact safely with delicate objects, 
making soft robotic structures promising for applications ranging from fruit harvesting \cite{wang2023development}, search and rescue \cite{dermaur2021RoBoa} to minimally invasive surgery \cite{wang2017cabledriven}.

In robotics, \textit{path planning} refers to the problem of finding a collision-free motion between an initial and a final configuration and identifying the required actuation while respecting the physical constraints of the system. Unlike \textit{trajectory planning}, which specifies movement over time, path planning is purely geometric: it seeks a continuous path in the configuration space (\cs{}) or workspace (\ws{}) of the robot without assigning timing.
For soft robots, path planning is particularly challenging. Their kinematics, i.e., their mappings between actuation in \cs{} and the resulting shapes in \ws{}, are highly nonlinear, and the number of degrees of freedom is effectively infinite. As a result, planning methods developed for rigid robots often perform poorly, and incorporating obstacle avoidance in cluttered environments remains computationally demanding.

\subsection{Related Work}

Path planning for soft robots has been approached from several directions, often borrowing techniques from rigid robotics but adapting them to the challenges of high-dimensional, continuously deformable bodies.

\textbf{Inverse kinematics-based methods.}
Many ``conventional'' path planning algorithms in rigid robotics rely on inverse kinematics: The desired robot or end-effector path is specified in \ws{}. Then, at each point along the path, the inverse kinematics are solved to find the necessary configuration in \cs{}, and these configurations are connected, possibly by interpolation, to form the continuous path.
Such methods have been explored in soft robotics, but their highly complex kinematics make them prone to geometric constraint violations, local minima, non-uniqueness, infeasibility of the solution, or high computational cost in cluttered environments \cite{neppalli2009ClosedForm, deng2019Nearoptimal, godage2015Modal}.
These drawbacks have motivated a shift towards methods that plan directly in \cs{}, avoiding explicit inverse kinematics.

\textbf{Artificial potential methods.}
Here, the robot is treated as a moving point in \cs{} that is drawn toward attractive potentials created by the goal configuration and repelled by obstacle potentials \cite{khatib1986RealTime}. The path emerges from following the gradient of this combined potential field. This approach has been validated computationally for curve-like obstacles \cite{fairchild2021Efficient} and experimentally for a cable-driven continuum robot \cite{ding2024Collisionfree}, but since these methods are based on continuous optimization, they can suffer from local minima.


\textbf{Sampling-based methods.}
Algorithms such as rapidly exploring random trees (RRT) are well established in rigid robotics for their efficiency and simplicity when planning in \cs{} \cite{lavalle2001Randomized}. Their extension RRT* adds asymptotic optimality \cite{karaman2011Samplingbased}.
For soft robots, the lack of closed-form inverse kinematics makes sampling challenging. Some works use relaxed kinematic constraints with collision checking \cite{bonilla2015Samplebased}; others combine RRT* with additional constraints, such as safety optimization \cite{luo2024Efficient}, \rev{or consider multitarget paths \cite{zhang2024cosine}.} 
Special efforts have also been made in the field of concentric tube robots for minimally invasive surgery, 
where RRTs explore \cs{} to find a path in \ws{} \cite{bergeles2013Planning, wu2015Motion, leibrandt2017Concentric}.
Alternatively, Jacobian-based inverse kinematics routines can project samples into feasible robot configurations \cite{meng2022RRTbaseda}.

\textbf{Graph-based methods.}
Also called roadmap techniques, these reduce the high-dimensional \cs{} to a one-dimensional, searchable path on a graph. Graphs can be built by connecting nearby configurations of a precomputed library ($k$-nearest neighbors), by randomly sampling and connecting collision-free states as in probabilistic roadmaps \cite{kavraki1998Analysis}, or by discretizing the space into a grid or motion primitives. Once built, the graph can be searched using algorithms such as Dijkstra \cite{dijkstra1959note}. It allows for custom edge costs such as geometric distance, actuation effort, or obstacle avoidance.
In soft robotics, examples include a time-augmented \cs{} graph for dynamic obstacle avoidance \cite{meng2021Anticipatory}, grid discretization for a soft bending pneumatic actuator \cite{shamilyan2024Resilient}, nearest-neighbor graphs for concentric tube robots \cite{niyaz2020Following}, and hybrid methods that generate configurations using inverse kinematics and then run Dijkstra’s search \cite{meng2021Smooth}.
Most of these methods still operate in \cs{}, even when the goal is specified in \ws{}.

\textbf{Obstacle representation.}
Obstacles can be incorporated into path planning through a variety of methods, e.g., intrinsically in artificial potential fields \cite{khatib1986RealTime}, explicit geometric intersection checks \cite{fairchild2021Efficient}, or volumetric occupancy maps. Among these, signed distance fields (SDFs) have become particularly popular for encoding the minimum distance from any point in the workspace to the nearest obstacle, enabling both fast collision detection and continuous clearance costs \cite{koptev2023Neural,li2024Representing}.
Although most SDFs are defined in \ws{}, they are emerging in \cs{} for high-dimensional planners \cite{li2024Configuration,long2025Neural}. In soft robotics, SDF-based models of continuum robot shapes can accelerate collision checks even in cluttered environments \cite{long2025Neurala}, where repeated geometric queries are often a major computational bottleneck \cite{fairchild2021Efficient}. \\[1.pt]

Fast and collision-free path planning for soft robots directly in \ws{} remains an open challenge: It requires generating sequences of robot shapes that are both collision-free and mechanically feasible. Existing methods either rely on approximate kinematic models or on computationally expensive inverse kinematics, which struggle with high degrees of freedom and cluttered environments. 

\subsection{Contributions}
To address the challenge of fast, collision-free path planning for soft robots, we propose a computationally efficient graph-based planning tool directly in \ws{}, or more appropriately, \emph{shape space} (\shs), which is better suited for soft robots due to their continuum nature.
Our method integrates a biomechanically exact forward kinematics model of a soft robotic arm \cite{kaczmarski2024Minimal} with a precomputed shape library and a $k$-nearest neighbor (\knn{}) graph that guarantees that all nodes correspond to physically valid shapes, enabling multi-objective planning at high speed.
By using signed distance fields (SDFs), clearance is precomputed and infeasible nodes are pruned before search, avoiding expensive collision checks during runtime.
This combination of exact modeling and offline graph construction allows us to eliminate dependence on inverse kinematics during planning and generate efficient and realistic paths. 
Our main contributions are as follows.
\begin{itemize}
\item \textbf{Kinematically exact biomechanical modeling for shape library}: Our shape library is generated from an elephant-trunk-inspired model whose forward kinematics are based on active filament theory, capturing coupled bending and twist from three fiber actuators with full mechanical accuracy, unlike simplified constant-curvature or reduced-order models. 
\item \textbf{Fast path planning directly in shape space}: Building a \knn{} graph from this library removes the need for inverse kinematics during planning and ensures that all planned shapes are realizable physical configurations.
\item \textbf{Multi-objective edge costs}: We incorporate geometric distance, actuation effort, and smoothness into the graph edge weights to simultaneously achieve obstacle avoidance and energy-efficient paths.
\end{itemize}
Due to the nature of our soft robotic arm, where fiber activations result in continuous shapes in the 3D space, we refer to the \textbf{activations space} (\acts{}) and \textbf{shape space} (\shs{}) instead of traditional configuration- or workspace representations throughout this work.

The letter is organized as follows: Section~\ref{sec:model} details the biomechanical soft robot model based on active filament theory. Section~\ref{sec:graph} describes the construction of the shape-space graph and path planning. In Section~\ref{sec:results}, we present simulation results validating our approach. Section~\ref{sec:discussion} discusses implications and limitations, and Section~\ref{sec:conclusion} concludes with future research directions.

\section{Biomechanical Soft Robot Model}\label{sec:model}
\begin{figure}
    \centering
    \includegraphics[width=0.8\columnwidth]{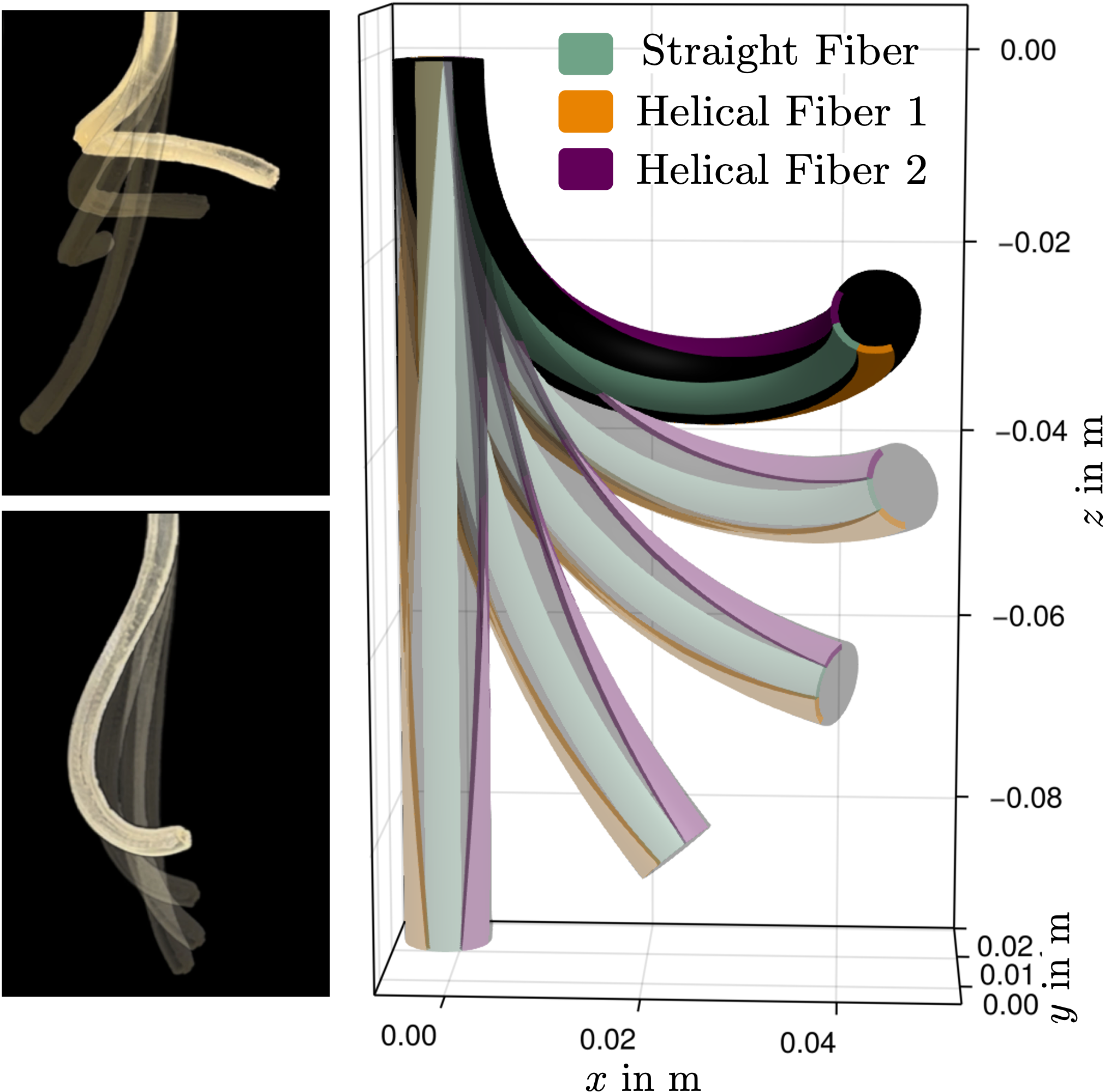}
    \caption{Three-fiber soft robotic arm \cite{kaczmarski2024Minimal, leanza2024Elephant}: The orientation of the fibers is inspired by the muscular structure of the elephant trunk, with one straight fiber and a helical fiber pair, enabling continuous deformations and great reachability.}
    \label{fig:trunk}
\end{figure}

Our soft robot is designed with three liquid crystal elastomer fibers \cite{leanza2024Elephant,kaczmarski2024Minimal}, 
inspired by the elephant trunk musculature (cf. Figure~\ref{fig:trunk}). 
Embedded in an elastic acrylate polymer cylinder, these three artificial muscle fibers can be selectively activated and achieve multimodal deformations that transition between modes for continuous movements.
By design, the structure is
\begin{itemize}
    \item {\it{slender}}: \rev{it is orders of magnitude longer than it is thick},
    \item {\it{soft}}: it can deform continuously, allowing for an infinite number of degrees of freedom,
    \item {\it{active}}: it can grow, shrink, or actively remodel its intrinsic shape,
\end{itemize}
which is why we model it as an \textit{active filament}. In contrast to passive filaments with fixed material and geometry properties, active filaments can actively contract, e.g., through magnetic actuation, biological growth, or muscular contraction. To describe such systems, we use the theory of morphoelasticity \cite{moulton2013Morphoelastic, kaczmarski2022Active}.
\rev{We note that Cosserat-rod based formulations play a central role in soft robotics that offer geometrically exact rod models and reduced-order strain parameterizations \cite{renda2020geometric}. The theory of morphoelasticity extends classical Cosserat-rod formulations by allowing for contraction and growth.}

\textbf{Morphoelastic filament theory.} A morphoelastic  filament is a slender structure that can grow, shrink, or actively remodel its intrinsic shape, while also deforming elastically. 
The theory considers a tubular body $\mathcal{B}_0 \in \mathbb{R}^3$, which is reduced to
\begin{itemize}
    \item a spatial curve $\boldsymbol{r}: [0, L] \rightarrow \mathbb{R}^3$, representing the centerline of the filament. The argument of the function is the material coordinate $\mechz$, and $L$ is the length of the filament in the reference configuration $\mathcal{B}_0$,
    \item an orthonormal director frame $\boldsymbol{D} = \{\boldsymbol{d}_1, \boldsymbol{d}_2, \boldsymbol{d}_3\}$, where $\boldsymbol{d}_i:[0,L] \rightarrow \mathbb{R}^3$, is attached to $\boldsymbol{r}(\mechz)$ for all $\mechz$.
\end{itemize}
The kinematics of the one-dimensional filament follow as
\begin{subequations}
\begin{align} \label{eq:main-ode}
    \bm{r}'(\mechz) &= \zeta \bm{d}_3,\\
    \bm{d}_i'(\mechz) &= \zeta \bm{u} \times \bm{d}_i, \quad i=\{1,2,3\},
    \label{eq:bvp-basic}
\end{align}
\end{subequations}%
with the axial extension of the filament $\zeta$ and the Darboux curvature vector $\bm{u} = u_1 \bm{d}_1 + u_2 \bm{d}_2 + u_3 \bm{d}_3$ describing the evolution of the director basis $\bm{D}$ along the filament \cite{moulton2013Morphoelastic}. We denote the properties of the filament in the absence of external loading as \emph{intrinsic} curvatures $\bm{u} = \bm{\hat u}$ and \emph{intrinsic} extension $\zeta = \hat \zeta$, \rev{i.e., the intrinsic shape is the shape induced by activation in the absence of external loads.}

Through the dimensionality reduction, we can represent the deformation $\boldsymbol{\chi}:\mathcal{B}_0 \rightarrow \mathcal{B}$ in terms of $\boldsymbol{r}$ and $\boldsymbol{D}$ as
\begin{align}
    \boldsymbol{\chi}(\boldsymbol{X}) = \boldsymbol{r}(\mechz) + \sum_{i=1}^3 \varepsilon e_i(\varepsilon R, \Theta, \mechz) \boldsymbol{d}_i(\mechz),
\end{align}
where $\boldsymbol{X}$ is a point in $\mathcal{B}_0$, $\{R, \Theta, \mechz\}$ are the cylindrical coordinates of $\boldsymbol{X}$, $\varepsilon$ is a small parameter of the thin rod geometry, and $e_i$ are the reactive strains that define the cross section deformation. The derivative of the deformation with respect to the undeformed coordinates defines the 
deformation gradient $\boldsymbol{F}=\nabla \boldsymbol{\chi}$ that we decompose multiplicatively, $\boldsymbol{F}=\boldsymbol{A} \boldsymbol{G}$, into an elastic tensor $\bm{A}$ and activation tensor $\bm{G}$.

\textbf{Active filament theory.} The active filament model represents a special case of morphoelasticity for filaments with fiber-based internal activation. It assumes that $\bm{G}$ satisfies the incompressibility constraint $\det(\bm{G})=1$. 
The active fiber architecture follows a fiber direction field $\bm{m}$ embedded in the filament body, which can be uniaxial or helical. In the absence of external loads, the intrinsic curvatures $\hat u_i$ and intrinsic extension $\hat \zeta$ of the robotic arm result from integrating the contribution of each fiber contraction across the cross section. While the proposed approach allows for any form of \emph{activation} $\gamma$, we consider the case of a piecewise uniform activation, such that we obtain one scalar value $\gamma_i \in \mathbb{R}$ per fiber. Negative activation corresponds to \emph{contraction}, positive activation corresponds to \emph{extension} \cite{kaczmarski2022Active}. 

\textbf{External loading.}
We define a deformed configuration $\mathcal{B}_\text{d}$ as a result of deforming the activated configuration $\mathcal{B}$ through external forces. We assume that the motion of the robotic arm is slow, and consider the quasi-static solution of the force and moment balance equations for morphoelastic rods \cite{moulton2013Morphoelastic},
\begin{align}
    \frac{\partial \bm{n}}{\partial \mechz} + \hat \zeta \bm{f} = \bm{0}, \quad
    \frac{\partial \bm{m}}{\partial \mechz} + \frac{\partial \bm{r}}{\partial \mechz} \times \bm{n} + \hat \zeta \bm{l} = \bm{0}.  
\end{align}\label{eq:bvp-extended}%
Here, $\bm{n}$ is the internal force, $\hat \zeta \bm{f}$ is the external body force per unit length of $\mathcal{B}_0$, $\bm{m}$ is the internal momentum, and $\hat \zeta \bm{l}$ is the external body couple per unit length of $\mathcal{B}_0$. The choices of $\bm{f}$ and $\bm{l}$ and the boundary conditions on $\bm{m}(\mechz)$ and $\bm{n}(\mechz)$ define different loading scenarios. 

\textbf{Three-fiber soft robotic arm.} In this work, we consider a structure with one longitudinal and two helical fibers (Figure~\ref{fig:trunk}) subject to gravitational load. Hence, its forward kinematics are governed by
\begin{subequations}  \label{eq:fk}
\begin{align}
    \bm{r}'(\mechz) &= \zeta \bm{d}_3,\\
    \bm{d}_i'(\mechz) &= \hat \zeta \bm{u} \times \bm{d}_i, \quad i=\{1,2,3\},\\
    \bm{0} &= \frac{\partial \bm{n}}{\partial \mechz} + \hat \zeta \bm{f}  \\
    \bm{0} &= \frac{\partial \bm{m}}{\partial \mechz} + \frac{\partial \bm{r}}{\partial \mechz} \times \bm{n} + \hat \zeta \bm{l}.
\end{align}
\end{subequations}
where the mapping from activation $\bm{\gamma}= [\gamma_1, \gamma_2, \gamma_3]$ to shape $\bm{r}(z)$ is mediated by the intrinsic curvature $\bm{\hat u}$ and extension $\hat \zeta$, both of which are functions of the fiber activations $\boldsymbol{\gamma}$, i.e. $\bm{\hat u}(\bm{\gamma})$ and $\hat \zeta(\bm{\gamma})$. The true curvature and extension under load are denoted $\bm{u}$ and $\zeta$. Details on the relationship between fiber geometry and intrinsic properties are discussed in \cite{kaczmarski2022Active}.

\section{Shape-Space Graph Construction and Search}\label{sec:graph}
Unlike optimization-based strategies, library-based approaches provide a robust alternative for inverse problems \cite{flaschel2026material}, as they avoid difficulties related to non-convex objective functions and local minima. Here, we adopt this concept in the context of path planning for the soft robot.
The biomechanically inspired model is the baseline for generating a library of physically feasible robot shapes that are then used to build a graph in \shs{} and find collision-free paths. A code example is provided on \href{https://github.com/cayobro/shape-space-graphs}{Github}.

\subsection{Shape Library Generation}
To generate the shape library, we solve the boundary value problem defined by \eqref{eq:fk}  to obtain the centerline $\bm{r}(z)$ of the three-fiber soft robotic arm  \cite{kaczmarski2024Minimal}. We use the  \textit{Julia} package developed for active filament theory \cite{kaczmarski2022Active}. As external loading, we consider gravitation due to the self-weight of the robot. We store each solution as a discrete centerline in a matrix $\bm{R} \in \mathbb{R}^{n_z \times 3}$ with its respective $(x,y,z)$-coordinates at $n_z$ locations between 0 and the robot length 
$L$, 
i.e. $R_{kj} = r_j(z_k)$, together with its activation vector $\boldsymbol{\gamma} = [\gamma_1, \gamma_2, \gamma_3] \in \mathbb{R}^3$. 
The collection of $N$ such solutions for random activations within the physical constraints $\boldsymbol{\gamma}\in [-\boldsymbol{\gamma}_\text{min}, \boldsymbol{\gamma}_\text{max}]$, where $\gamma_{\text{min},i}=-1.67$, $\gamma_{\text{max},i}=0$ for $i=1,2,3$, to only allow for contraction of the fiber, not extension, defines our shape library
\begin{align}
\mathcal{R} = \{ (\bm{\gamma}^{(i)},\bm{R}^{(i)}) \}_{i=1}^N
\end{align}
with activation-shape tuples $(\bm{\gamma}^{(i)},\bm{R}^{(i)})$.

\subsection{$k$-Nearest Neighbor Graph in Shape Space}
From the shape library, we construct an undirected \knn{} graph $G=(V,E)$.
Each node or vertex $i \in V$ corresponds to an activation-shape tuple $(\bm{\gamma}^{(i)},\bm{R}^{(i)})$.
We use a root-mean-square metric to measure the similarity between two shapes in \shs{},  
\begin{align}
d_{\text{RMS}}(\bm{R}^{(i)},\bm{R}^{(j)}) = \sqrt{ \tfrac{1}{n_z} \sum_{k=1}^{n_z} \sum_{j=1}^{3} \left( R_{kj}^{(i)} - R_{kj}^{(j)} \right)^2 },\label{eq:d_rms}
\end{align}
where the sum is taken over the $n_z$ discrete centerline points. 
Based on this, we identify the $k$ geometrically closest shapes of the shape matrix $\bm{R}^{(i)}$ for every node $i$ by forming edges
\begin{align}
E = \{ (i,j) \mid i \in \mathcal{N}_k(j) \text{ or } j \in \mathcal{N}_k(i) \}. \label{eq:edges}
\end{align}
Here, $\mathcal{N}_k(i)$ denotes the neighborhood of each node $i$.
This results in a sparse roadmap where each node is guaranteed to be a physically realizable configuration of the robot.
We deliberately use this simple distance for our neighbor search since it is robust, scales well, and finds neighbors fast. We will elaborate more meaningful multi-objective edge costs, e.g., to include activation magnitude, in the next step. This two-stage strategy avoids the $\mathcal{O}(N^2)$ cost of evaluating custom edge weights between all pairs of nodes.

\subsection{Collision Avoidance with Signed Distance Fields}

To avoid collisions with the robot's environment, we use signed distance fields (SDFs) to identify and remove nodes and edges from the graph that would intersect with obstacles in the environment.

An SDF $\phi:\mathbb{R}^3\!\to\!\mathbb{R}$ encodes for any point $\boldsymbol{p}\in \mathbb{R}^3$ the orthogonal distance to the closest boundary $\partial \Omega$ of an obstacle $\Omega \in  \mathbb{R}^3$ in \shs{}, i.e.,
\begin{align}
    \phi(\boldsymbol{p}) = \begin{cases}
        + \min_{\boldsymbol{q}\in\partial\Omega} \Vert \boldsymbol{p} - \boldsymbol{q}\Vert, \ \boldsymbol{{p}} \not\in \Omega \\
        - \min_{\boldsymbol{q}\in\partial\Omega} \Vert \boldsymbol{p} - \boldsymbol{q}\Vert, \ \boldsymbol{{p}} \in \Omega
    \end{cases}
\end{align}
such that 
$\phi(\mathbf{p})>0$ outside obstacles, $\phi(\mathbf{p})=0$ on their surface, and $\phi(\mathbf{p})<0$ inside. We here focus on relatively simple shapes, including cuboids and cylinders, for which solving the minimization problem $\min_{\boldsymbol{q}\in\partial\Omega} \Vert \boldsymbol{p} - \boldsymbol{q}\Vert$ is trivial and allows for a closed-form implementation. However, SDFs for more complex shapes may be implemented by solving the minimization problem numerically.

For a centerline shape $\bm{R}^{(i)}\in \mathbb{R}^{n_z\times 3}$ consisting of $n_z$ discrete points along the robot body, we define the node clearance as the minimum signed distance along the entire centerline $z$
\begin{align}
    c^{(i)} = \min_{k=1,...,n_z} \phi\left(R_{k1}^{(i)},R_{k2}^{(i)},R_{k3}^{(i)}\right) - \rho_\text{tube},
\end{align}
where $\rho_\text{tube}>0$ is the radius of the tube-shaped soft robotic arm and
$R_{k1}^{(i)},R_{k2}^{(i)},R_{k3}^{(i)}$ are the respective $(x,y,z)$-coordinates of the $i$-th centerline. This formulation ensures that the entire centerline, and not just the tip, is considered.
For the more general case, when multiple obstacles are defined through their SDFs $\phi_j$ with $j=1,...,n_o$, we define the node clearance as
\begin{align}
    c^{(i)} = \min_{j=1,...,n_o} \min_{k=1,...,n_z} \phi_j\left(R_{k1}^{(i)},R_{k2}^{(i)},R_{k3}^{(i)}\right) - \rho_\text{tube},
\end{align}
that is, the total node clearance is the minimum of the node clearances of the individual obstacles.
In practice, we precompute the node clearance $\{c^{(i)}\}$ offline using analytic obstacle SDFs, e.g., cylinders, boxes.
We retain nodes with positive clearance $c>0$ and prune all nodes with $c\leq 0$. Finally, we perform edge sweeps by evaluating $\phi$ along linear interpolants between nodes, pruning edges that intersect obstacles. \rev{While we focus on simple analytical obstacles, the SDF-based pruning is not restricted to simple geometries, but works for any obstacle for which we can evaluate an SDF, including non-convex or mesh-based obstacles.}

\subsection{Edge Weight Formulations}
Assigning edge weights to the resulting candidate edges is a key feature of graph-based path planning, as this will define the shortest path, in terms of lowest cost, from a start node to an end node. \rev{It also} allows us to constrain certain paths and prefer others. With our shape library consisting of tuples $(\bm{\gamma}^{(i)},\bm{R}^{(i)})$ of activations and centerline shapes, we can formulate edge costs both in \acts{} and in \shs{}.  

\textbf{Geometric distance.} Since geometric proximity is our main priority, we include $d_{\text{RMS}}(\bm{R}^{(i)},\bm{R}^{(j)})$ from \eqref{eq:d_rms} as part of the edge costs. However, this metric alone does not reflect activation energy or smoothness, and may result in energetically costly or “bumpy” paths in \acts{}.  

\textbf{Activation energy.}  
To account for the effort of the \rev{actuators}, we penalize large absolute activation levels through 
the edge weight, defined in \acts{},  
\begin{align}
d_\text{mag}(\bm{\gamma}^{(i)}, \bm{\gamma}^{(j)}) = \tfrac{1}{2} \left( \bm{\gamma}^{(i)\top} \bm{K} \bm{\gamma}^{(i)} + \bm{\gamma}^{(j)\top} \bm{K} \bm{\gamma}^{(j)}\right),
\end{align}
where $\bm{K}$ is a weighting matrix that can rescale or weight the contributions of individual fibers.  
For our soft robot, $\bm{K}=\bm{I}$.

\textbf{Activation rate.}  
In addition to keeping the activation magnitude low, it is desirable to reduce rapid heating and cooling ramps \rev{for actuators such as} liquid crystal elastomer fibers. Hence, we penalize large activation jumps between consecutive activation vectors, 
\begin{align}
d_\text{smo}(\bm{\gamma}^{(i)}, \bm{\gamma}^{(j)}) = \lVert \bm{\gamma}^{(i)}-\bm{\gamma}^{(j)} \rVert^2,
\end{align}
which results in smooth and gradual changes in the activations along the path.  

\textbf{Multi-Objective Edge Costs.}
Last but not least, the modular edge costs representing geometric distance $d_\text{RMS}(\bm{R}^{(i)},\bm{R}^{(j)})$, activation magnitude $d_\text{mag}(\bm{\gamma}_i,\bm{\gamma}_j)$, and smoothness $d_\text{smo}(\bm{\gamma}_i,\bm{\gamma}_j)$ are combined into a multi-objective edge cost in both \shs{} and \acts{} as
\begin{align}
    w(i,j) = &\alpha \, d_\text{RMS}(\bm{R}^{(i)},\bm{R}^{(j)}) \notag \\
    & \quad + \beta \, d_\text{mag}(\bm{\gamma}^{(i)},\bm{\gamma}^{(j)}) 
    + \delta \, d_\text{smo} (\bm{\gamma}^{(i)},\bm{\gamma}^{(j)}) \label{eq:multi-obj-cost}
\end{align}
with weighting parameters $\alpha>0$, $\beta\geq 0$, $\delta\geq0$. They represent the trade-off between the different objectives.

\subsection{Graph Search for Path Planning}
With the \knn{} graph in which each edge has assigned weights $w(i,j)$, path planning reduces to finding the path with the minimum cost between start and end nodes. For this purpose, we use Dijkstra's algorithm \cite{dijkstra1959note}, which guarantees finding the optimal path in terms of the cumulative edge costs.
In each step, the algorithm picks the unvisited node with the lowest cost, then calculates the distance through it to each unvisited neighbor to check whether a cheaper path to its neighbors exists. Then, it updates the costs accordingly. The algorithm runs until it reaches the end node.
Since all edge costs are non-negative, Dijkstra’s algorithm is guaranteed to converge to the global optimum without requiring any heuristics. This makes it a natural baseline for our framework and, with our sparse graph, planning typically completes in the order of milliseconds.
The resulting planned path is a sequence of nodes
\begin{align}
    \pi = (i_0, i_1, \dots, i_m), \qquad i_k \in V,
\end{align}
where $i_0$ and $i_m$ denote the start and goal nodes and consecutive pairs $(i_k,i_{k+1})$ are connected by edges $(i_k,i_{k+1}) \in E$. \rev{To obtain a continuous actuation over time, we can interpolate within these discrete sequences.}

\begin{figure*}
    \centering
    \includegraphics[width=\textwidth]{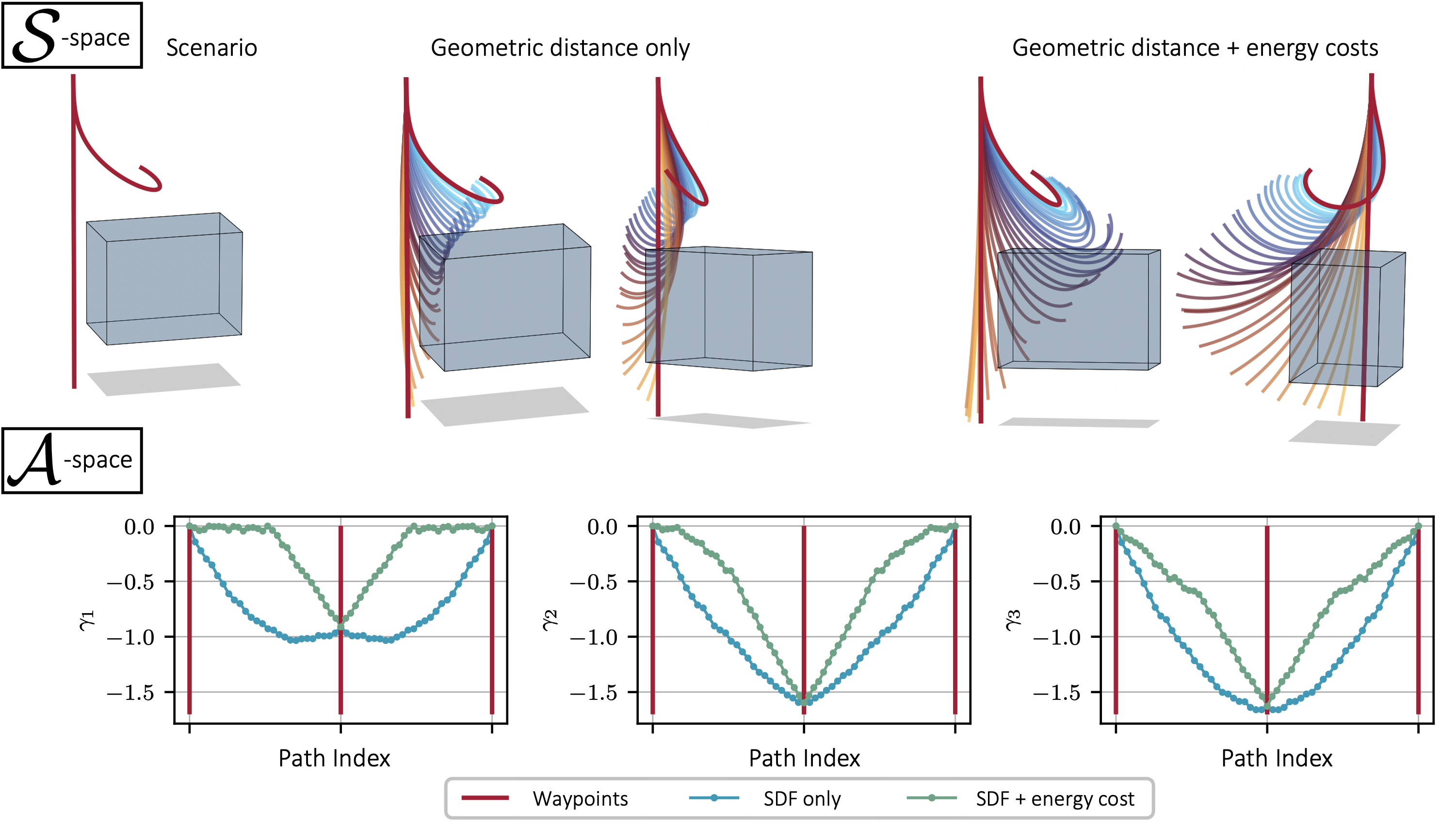}
    \caption{\textbf{Sequence of shapes $\bm{r}(z)$ (top) and activations $\bm{\gamma}$ (bottom) of the box obstacle scenario} for two different edge cost formulations. Shapes are color-coded according to progression along the robot’s path segments. Activations $\gamma_i$ are depicted for each fiber to compare in between the edge cost formulations. }
    \label{fig:box}
\end{figure*}

\begin{figure*}
    \centering
    \includegraphics[width=\textwidth]{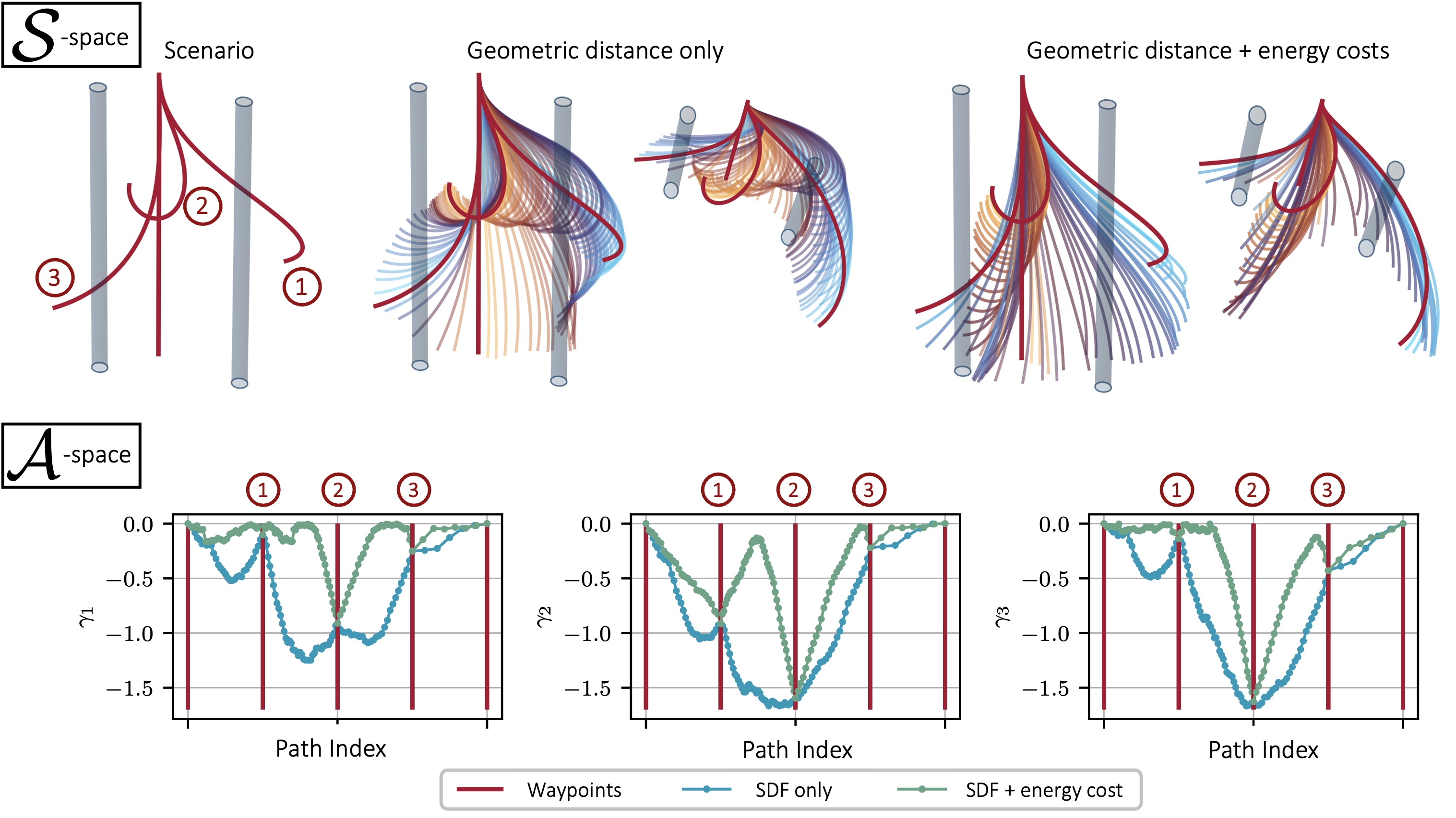}
    \caption{\textbf{Sequence of shapes $\bm{r}(z)$ (top) and activations $\bm{\gamma}$ (bottom) of the cylinder obstacle scenario} with multiple waypoints and for two different edge cost formulations. Shapes are color-coded according to progression along the robot’s path segments. Activations $\gamma_i$ are depicted for each fiber to compare in between the edge cost formulations, for normed path segments.
    }
    \label{fig:cylinder}
\end{figure*}

\section{Results}\label{sec:results}
We evaluate the proposed planning framework with the precomputed shape library of the soft robotic arm that contains $N=100,000$ samples with a spatial discretization of $n_z=100$. We build the \knn{} graph with $k=20$ neighbors. 
\rev{The fiber architecture follows a minimal design with one longitudinal and two symmetrical helical fibers as the minimum number of actuators that  reproduce key deformations of the elephant trunk \cite{kaczmarski2024Minimal, leanza2024Elephant}. This design was validated experimentally on the already mentioned three-fiber prototype, where the two helical fibers complete $\Omega=\qty{\pm 108}{\degree}$ of revolution along the trunk length, and the corresponding local helix angles are defined by $\alpha=\arctan({R}/{L} \; \Omega)$ where ${R}/{L}= {1}/{20}$ is the ratio between length $L=\qty{0.09}{\meter}$ and radius $R$.}

We compare two scenarios: a single-waypoint route with a bulky obstacle obstructing the path, and a multi-waypoint route requiring navigation through two narrow cylinders. To emphasize the impact of the additional edge costs formulated in \acts{}, we compare for both scenarios the geometric-distance-only edge cost ($\alpha=1$, $\beta=\delta=0$)
with the multi-objective edge costs including energy terms ($\alpha=\beta=\delta=1$) as defined in~\eqref{eq:multi-obj-cost}. 
To evaluate the edge weight formulations, we evaluate a path consisting of $m$ nodes with the metrics
tip path length
\begin{align}
    L_{\text{tip}}(\pi) = \sum_{k=0}^{m-1} \sqrt{\sum_{j=1}^{3}\left(R_{n_z j}^{{(i_{k+1})}} - R_{n_z j}^{(i_k)} \right)^2 },
\end{align}
the quadratic activation effort 
\begin{align}
    E_\gamma (\pi) = \sum_{k=0}^{m} \lVert \bm{\gamma}^{(i_k)} \rVert^2,
\end{align}
and the activation smoothness 
\begin{align}
    \mathrm{TV}_\gamma(\pi) = \sum_{k=0}^{m-1} \lVert \bm{\gamma}^{(i_{k+1})} - \bm{\gamma}^{(i_k)} \rVert.
\end{align}
Summary metrics for both scenarios are presented in Table~\ref{tab:metrics}.
\rev{We performed all simulations on a standard Apple M3 Pro with 18~GB of memory. The shape library with 100,000 samples occupies \qty{230.4}{\mega\byte}, and its generation takes \qty{180.67}{\second}.}

\subsection{Bulky Obstacle in the Way}
First, we explore a scenario where the soft robot rises from the zero-position, or steady state, to a curled-up shape and back, with a bulky, box-shaped obstacle in the path.
Figure~\ref{fig:box} highlights the scenario, showing the sequences of activations $\bm{\gamma}$ and shapes $\bm{r}(z)$ for both edge cost formulations. 
Adding energy costs drastically reduces the activation effort of the fibers by almost \qty{50}{\percent}, while still reliably avoiding the obstacle. However, this comes at the cost of a \qty{73}{\percent} longer tip trajectory and slightly more spatially expanded nodes, suggesting the search must balance competing cost terms more intensively.
Strikingly, including energy costs results in more elongated shapes: Since activation effort corresponds to \textit{contraction}, less contracted shapes are more energy efficient.
Interestingly, the combined inclusion of the smoothness term and energy costs does not decrease activation variability; indeed, the activation smoothness metric $\mathrm{TV}_\gamma$ slightly increases. This arises because prioritizing minimal activation energy drives the algorithm to select states with low activations but requires steeper transitions between them. In practice, the relative weighting of energy and smoothness costs can be tuned depending on application requirements.

\subsection{Multi-Waypoint Route Through Cylindrical Obstacles}
In the second scenario, the soft robot navigates between two cylindrical obstacles and reaches multiple shapes along the route, beginning and ending in the zero-position. Such scenarios are motivated by grippers needing to place objects in various locations while avoiding obstacles like shelves or machinery. 
\rev{We select waypoints as shapes from the precomputed library, place them close to obstacles to increase difficulty, and plan the motion as a sequence of local start–goal problems between two consecutive waypoints.}

Figure~\ref{fig:cylinder} illustrates the scenario with sequences of activations $\bm{\gamma}$ and shapes $\bm{r}(z)$.
Notably, the algorithm finds a path quickly through all five waypoints and successfully avoids collisions. As in the box scenario, we observe a trade-off: the tip path length increases by \qty{77.78}{\percent}, while energy costs decrease by \qty{69.12}{\percent} when energy terms are included in the edge cost formulation. The reduced graph search computation time relative to the box scenario may reflect non-deterministic variations in search efficiency.
Consistent with the first scenario, the activation smoothness metric $\mathrm{TV}_\gamma$ slightly increases with energy cost inclusion. This reflects the inherent trade-off between minimizing activation energy and maintaining smooth activation transitions when both smoothness and energy terms are combined.

\begin{table}
  \centering
  \begin{minipage}{0.2\columnwidth}
    \includegraphics[width=\textwidth]{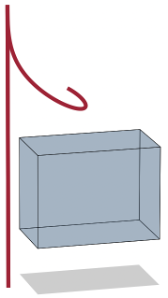}
  \end{minipage} 
  \begin{minipage}{0.75\columnwidth}
    \centering
    \begin{tabular}{cccc}
     & SDF only & SDF + energy  & \\
    \toprule
    $n_\text{nodes}$ & 55 & 63 & {\color{palo}{+}\qty{14.55}{\percent}} \\ \midrule
    $L_\text{tip}$ & \qty{0.15}{\meter} & \qty{0.26}{\meter} & {\color{palo}{+}\qty{73.33}{\percent}} \\ \midrule
    $E_\gamma$ & 172.76 & 87.23 & {\color{cardinal}{-}\qty{49.49}{\percent}} \\ \midrule
    $TV_\gamma$ & 5.39 & 5.56 & {\color{palo}{+}\qty{3.15}{\percent}} \\ \midrule
    Search & \qty{0.480}{\second} & \qty{0.424}{\second} & {\color{cardinal}{-}\qty{11.67}{\percent}} \\
    \bottomrule
    \end{tabular}
  \end{minipage} %
    \\[5mm]
    \begin{minipage}{0.2\columnwidth}
    \includegraphics[width=\textwidth]{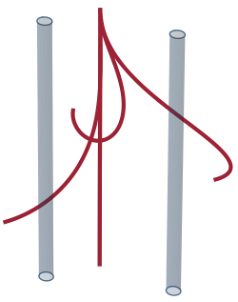}
  \end{minipage} 
  \begin{minipage}{0.75\columnwidth}
    \begin{tabular}{cccc}
     & SDF only & SDF + energy  & \\
    \toprule
    $n_\text{nodes}$ & 108 & 110 & {\color{palo}{+}\qty{1.85}{\percent}} \\ \midrule
    $L_\text{tip}$ & \qty{0.27}{\meter} & \qty{0.48}{\meter} & {\color{palo}{+}\qty{77.78}{\percent}} \\ \midrule
    $E_\gamma$ & 318.51 & 98.34 & {\color{cardinal}{-}\qty{69.12}{\percent}} \\ \midrule
    $TV_\gamma$ & 7.57 & 8.39 & {\color{palo}{+}\qty{10.83}{\percent}} \\ \midrule
    Search & \qty{0.695}{\second} & \qty{0.491}{\second} & {\color{cardinal}{-}\qty{29.35}{\percent}} \\
    \bottomrule
    \end{tabular}
    \end{minipage}
    \caption{Metrics from the paths where obstacle avoidance is combined with energy efficiency. Note that normalized activation values without units are reported.}
    \label{tab:metrics}
\end{table}

\mycomment{\subsection{Effect of shape-space database}
\NOTE{Is this useful? If not, we can also kick out the extensive introduction of the metrics at the beginning of Section 4.}

Only use geometric distance edge weight: $\alpha=1$, $\beta=\delta=\lambda=0$. 
\begin{itemize}
	\item shape discretization: Compare r with 50, 100, 200 points. 
	- Metrics: path length, computation time.
	- Visualize: table of metrics + convergence plot.
	\item library size: Compare 5k, 10k, 20k shapes. 
	- Metrics: T, number of visited nodes.
	- Visualize: line plot (runtime vs. library size).
\end{itemize}
Somehow also include variation in $k$ parameter of graph!!!

\begin{table}[h]
	\centering
	\caption{Effect of shape discretization.}
	\label{tab:placeholder}
	\begin{tabular}{c|ccc}
		\toprule
		$z$-discretization $n_z$ & $L$ & $n_\text{s}$ & $T$ \\
		\midrule
		10 & & & \\
		20 & & & \\
		50 & & & \\
		100 & & & \\
		200 & & & \\
	\end{tabular}
\end{table}

\begin{table}[h]
	\centering
	\caption{Effect of shape library size.}
	\label{tab:placeholder}
	\begin{tabular}{c|ccc}
		\toprule
		Library size $N$ & $L$ & $n_\text{s}$ & $T$ \\
		\midrule
		1,000 & & & \\
		10,000 & & & \\
		100,000 & & & \\
	\end{tabular}
\end{table}
}

\section{Discussion}\label{sec:discussion}
This work introduces a graph-based path planning tool for soft robotic arms that ensures physically valid configurations by coupling a biomechanical model with a precomputed shape-space graph. The \knn{} graph, constructed with $k=20$ neighbors from a shape library of 100{,}000 samples and multi-objective edge costs, quickly identifies collision-free, energy-aware paths.
Our results suggest that incorporating energy costs naturally leads to trajectories that trade longer paths for reduced activation effort, because elongated fibers are less energy-demanding. 


\rev{\textbf{Limitations.}}
\emph{Scalability} remains a challenge: $100{,}000$ samples provide a good trade-off between smoothness and memory/computation time, and larger libraries may need compression techniques such as encoder–decoder architectures to remain practical. 
\rev{Still, building the graph takes up to several minutes, limiting the approach to \emph{static or infrequently changing obstacles}. For more dynamic environments, incremental updates of the graph need to be explored. 
To include \emph{task-specific applications}, the edge costs can be augmented with penalties on deviations from desired end-effector paths in workspace. Generally, a systematic exploration of parameters associated with generating the graph and costs is necessary.  
Finally, all results are computational, and \emph{validation on hardware}
is an important next step. Limitations coupled to the experimental validation include that 
(i) the method plans \emph{quasi-static sequences} without enforcing dynamic feasibility or actuator-rate constraints, so continuous execution on hardware might require coupling to an appropriate dynamic controller, and that
(ii) the shape library is generated for a fixed, history-independent activation–shape relation under given loading: cyclic effects and hysteresis in real actuators may shift the reachable shapes over time and would then require updating or adapting the library. 
We envision using the same three-fiber LCE prototype as developed previously \cite{leanza2024Elephant}, where each fiber contains an embedded resistive wire and has its own electrical channel, so the three fibers can be Joule-heated and controlled independently to generate different deformation modes. Alternatively, a similar fiber architecture could be realized with tendon- or cable-driven robots, where cables routed in the same longitudinal and helical layout are actuated by motors to reproduce the same bending and twisting capabilities. 
}

\section{Conclusion}\label{sec:conclusion}
This study presents a fast, energy-aware, collision-free path planning tool for soft robots with biomechanical fidelity inspired by elephant trunks. 
Importantly, this approach is generalizable and applicable to any soft robot for which an accurate, yet computationally costly, model exists, enabling practical planning even when real-time model evaluation is infeasible.
Looking ahead, extending static planning to dynamic closed-loop control that integrates actuation, sensing, and feedback presents challenges due to the high-dimensional and nonlinear dynamics of soft robots.

\bibliography{SoftRobots.bib}
\bibliographystyle{IEEEtran}

\end{document}

%% file: defs.tex
\newcommand{\cs}{$\mathcal{C}$-space}
\newcommand{\ws}{$\mathcal{W}$-space}
\newcommand{\shs}{$\mathcal{S}$-space}
\newcommand{\acts}{$\mathcal{A}$-space}
\newcommand{\knn}{$k$-NN}
\newcommand{\mechz}{z}
\definecolor{dunkelblau}{rgb}{0.0, 0.2314, 0.6196}%
\definecolor{hellblau}{rgb}{0.000, 0.7451, 1.0000}%
\definecolor{rot}{rgb}{0.6980,0.1333,0.1333}%
\definecolor{cardinal}{rgb}{0.6156, 0.1333, 0.2078}
\definecolor{palo}{rgb}{0, 0.4157, 0.3216}

\newcommand{\NOTE}[1]{\textcolor{blue}{#1}}

\newcommand{\mycomment}[1]{}

\newcommand{\rev}[1]{\textcolor{black}{#1}}
\newlength{\mlLegendThickness}
\newlength{\mlLegendHeight}